\documentclass[runningheads]{llncs}
\usepackage[T1]{fontenc}
\usepackage{inconsolata}  
\usepackage{xcolor}
\usepackage[T1]{fontenc}  

\usepackage{amsmath}
\usepackage{amssymb}
\usepackage{bm}
\usepackage{bbm}

\usepackage{tabularx}
\usepackage{multirow}
\usepackage{makecell}
\usepackage{booktabs}
\usepackage{colortbl}  
\usepackage{array}  
\usepackage{marvosym}
\usepackage{graphicx}  
\usepackage{overpic}  
\usepackage{verbatim}  
\usepackage{enumitem}  
\usepackage{adjustbox}  
\usepackage{minitoc}  
\usepackage[accsupp]{axessibility}  

\usepackage{pifont}  
\usepackage{listings}
\usepackage{textcomp}
\definecolor{ForestGreen}{rgb}{0.13, 0.55, 0.13}  
\definecolor{LGray}{gray}{0.95}  
\definecolor{red}{rgb}{1.0, 0.0, 0.0}  
\definecolor{violet}{rgb}{0.56, 0.0, 1.0}  

\usepackage[colorlinks, linkcolor=blue, citecolor=blue]{hyperref}
\usepackage{makecell}
\usepackage{ifsym}
\begin{document}
\title{MedSeg-R: Reasoning Segmentation in Medical Images with Multimodal Large Language Models}
%

\author{Yu Huang\inst{1}$^\dag$ 
\and Zelin Peng\inst{1}$^\dag$ 
\and Yichen Zhao\inst{1}
\and Piao Yang\inst{2}
\and \\ Xiaokang Yang\inst{1} 
\and Wei Shen$^{1(\textrm{\Letter})}$ 
}

\institute{$^1$ MoE Key Lab of Artificial Intelligence, AI Institute, \\Shanghai Jiao Tong University, Shanghai, China. \\
$^2$ Department of Radiology, The First Affiliated Hospital, \\ Zhejiang University School of Medicine, Hangzhou, Zhejiang, China }

\maketitle    

\newcommand\blfootnote[1]{%
\begingroup 
\renewcommand\thefootnote{}\footnote{#1}%
\addtocounter{footnote}{-1}%
\endgroup 
}
\blfootnote{$^{\textrm{\Letter}}$ Corresponding Author: \texttt{wei.shen@sjtu.edu.cn}}
\blfootnote{$^\dag$ Indicates equal contribution.}

\begin{abstract}
Medical image segmentation is crucial for clinical diagnosis, yet existing models are limited by their reliance on explicit human instructions and lack the active reasoning capabilities to understand complex clinical questions. While recent advancements in multimodal large language models (MLLMs) have improved medical question-answering (QA) tasks, most methods struggle to generate precise segmentation masks, limiting their application in automatic medical diagnosis. In this paper, we introduce medical image reasoning segmentation, a novel task that aims to generate segmentation masks based on complex and implicit medical instructions. To address this, we propose MedSeg-R, an end-to-end framework that leverages the reasoning abilities of MLLMs to interpret clinical questions while also capable of producing corresponding precise segmentation masks for medical images. It is built on two core components: 1) a global context understanding module that interprets images and comprehends complex medical instructions to generate multi-modal intermediate tokens, and 2) a pixel-level grounding module that decodes these tokens to produce precise segmentation masks and textual responses. Furthermore, we introduce MedSeg-QA, a large-scale dataset tailored for the medical image reasoning segmentation task. It includes over 10,000 image-mask pairs and multi-turn conversations, automatically annotated using large language models and refined through physician reviews. Experiments show MedSeg-R's superior performance across several benchmarks, achieving high segmentation accuracy and enabling interpretable textual analysis of medical images.

\end{abstract}
%
%
%


\section{Introduction}

\begin{figure}[t]
    \centering
    \includegraphics[width=1.0\linewidth]{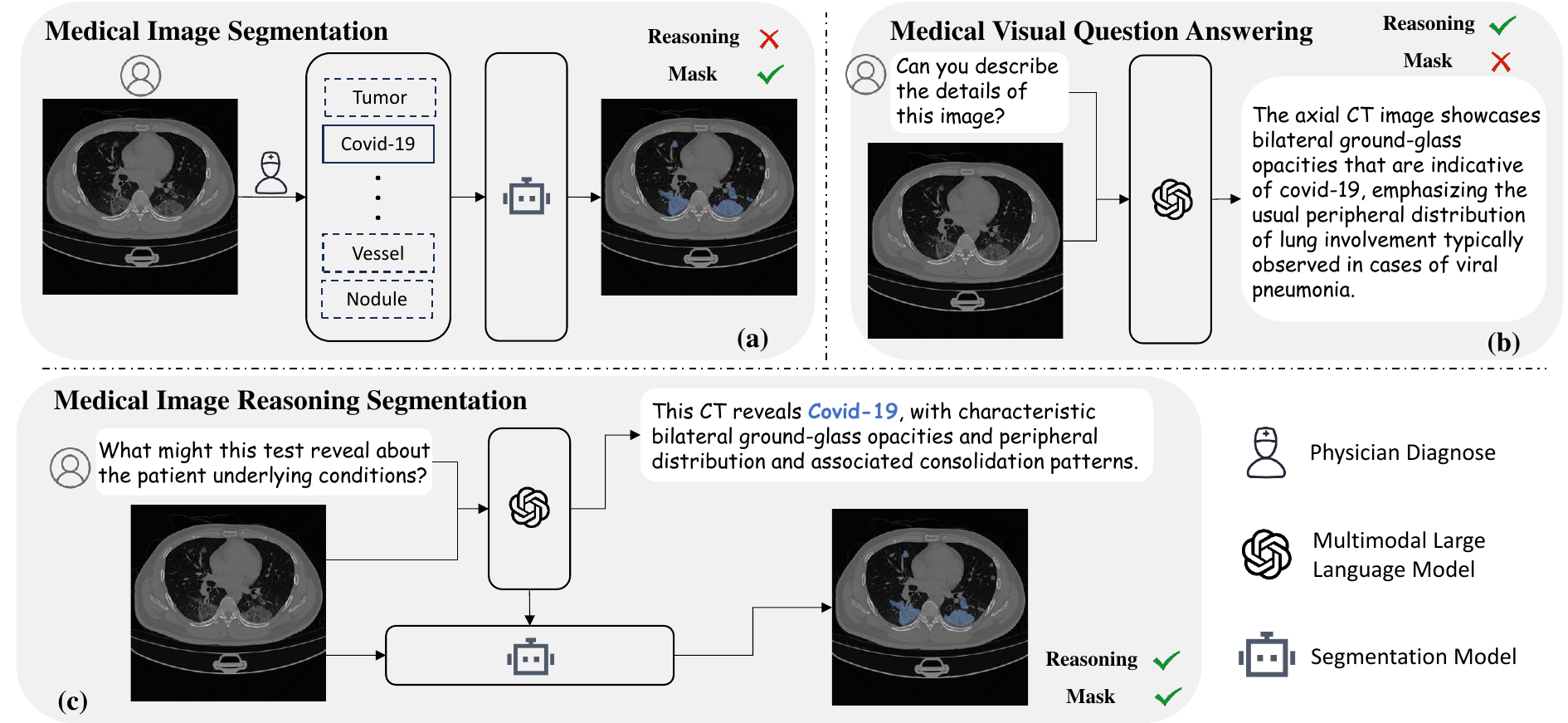}
    \caption{\textbf{Different tasks in medical image analysis.} The medical image segmentation task (a) heavily relies on explicit human instructions to segment the desired class, while the medical visual question-answering task lacks pixel-level grounding capabilities. Our novel task, medical image reasoning segmentation (c), requires the model to simultaneously generate text response and corresponding segmentation masks.}
    \label{fig1}
    \vspace{-4mm}
\end{figure}

Medical image segmentation plays a crucial role in clinical diagnosis and real-world medical decision-making. Recent segmentation models \cite{MedSAM,biomedparse,samed,nnunet,samparser,parameterefficientfinetuning} have demonstrated high accuracy in delineating various anatomical structures and pathological regions. However, while these models excel at segmenting images within predefined categories, they still rely heavily on explicit instructions from humans, such as ``\texttt{Identify the COVID-19 infected area,}'' which provides a straightforward reference like ``\texttt{COVID-19.}'' In an ideal autonomous medical diagnosis system, physicians would pose more open-ended queries, such as ``\texttt{What possible conditions are indicated by this examination?}'' In response, the system needs to provide a detailed description of the patient's condition along with the corresponding segmentation results (see Fig.~\ref{fig1}(a)). However, current models often lack such a reasoning ability, which presents a significant challenge when addressing the complex and variable nature of clinical inquiries.


Recently, the rapid advancement of multimodal large language models (MLLMs) \cite{biomedgpt,pubmedclip,biomedparse,llavamed,glamm} in the medical field has opened new avenues for research and development. These models exhibit exceptional capabilities in understanding and processing complex visual-language instructions and leveraging advanced reasoning mechanisms to enhance downstream tasks. Consequently, MLLMs have demonstrated remarkable performance in various vision-language applications, including biomedical visual question answering (VQA) and image captioning (as shown in Fig.~\ref{fig1}(b)). However, as shown in Table \ref{tab:methods_comparison}, most of these models are designed to generate text response and thus lack pixel-level grounding capabilities. Although a few segmentation models with reasoning ability, such as LISA \cite{lai2023lisa}, have been proposed, they still primarily produce text responses, such as ``\texttt{Sure, it is [SEG].}\footnote{The token \texttt{[SEG]} serves as a placeholder to instructs the model to generate segmentation masks.}'' This indicates that the potential of harnessing MLLMs' reasoning ability for pixel-level grounding tasks, such as medical image segmentation, remains largely unexplored.

\begin{table*}[!t]
    \footnotesize
    \centering
    \tabcolsep=0.12cm
    \renewcommand{\arraystretch}{1.2}
    \caption{\textbf{Comparison of Recent Medical Large Multimodal Models (LMMs).}
    The \textit{pixel-level grounding} column highlights models capable of generating segmentation masks, while \textit{multi-round Conversation} indicates models that support interactive dialogues with users. Our proposed model distinguishes itself by integrating both pixel-wise grounding and conversational capabilities within an end-to-end training framework, enabling more comprehensive and adaptable medical image analysis.}
    \begin{tabular}{lccc}
        \toprule
        Method & \makecell{Pixel-level \\ Grounding} & \makecell{Multi-round \\ Conversations} & \makecell{End-End \\ Model} \\
        \midrule

        \rowcolor{LGray} miniGPT4 (arXiv-23)~\cite{zhu2023minigpt} & \textcolor{ForestGreen}{\ding{51}} & \textcolor{ForestGreen}{\ding{51}} & \textcolor{ForestGreen}{\ding{51}} \\

        PubMedCLIP (ACL-2023)~\cite{pubmedclip} & \textcolor{red}{\ding{55}} & \textcolor{ForestGreen}{\ding{51}} & \textcolor{ForestGreen}{\ding{51}} \\

        \rowcolor{LGray} Instruct-BLIP (arXiv-23)~\cite{dai2023instructblip} & \textcolor{red}{\ding{55}} & \textcolor{ForestGreen}{\ding{51}} & \textcolor{ForestGreen}{\ding{51}} \\

        \arrayrulecolor{black}

        BiomedGPT (Nat Med 2024)~\cite{biomedgpt} & \textcolor{red}{\ding{55}} & \textcolor{ForestGreen}{\ding{51}} & \textcolor{ForestGreen}{\ding{51}} \\

        \rowcolor{LGray} BiomedCLIP (arXiv-23)~\cite{zhang2023biomedclip} & \textcolor{red}{\ding{55}} & \textcolor{ForestGreen}{\ding{51}} & \textcolor{red}{\ding{55}} \\

        LISA (CVPR 2024)~\cite{lai2023lisa} & \textcolor{ForestGreen}{\ding{51}} & \textcolor{red}{\ding{55}} & \textcolor{ForestGreen}{\ding{51}} \\

        \rowcolor{LGray} BiomedParse (Nat Methods 2024)~\cite{biomedparse} & \textcolor{ForestGreen}{\ding{51}} & \textcolor{red}{\ding{55}} & \textcolor{ForestGreen}{\ding{51}} \\

        LLaVA-Med (NeurIPS-23)~\cite{llavamed} & \textcolor{red}{\ding{55}} & \textcolor{ForestGreen}{\ding{51}} & \textcolor{ForestGreen}{\ding{51}} \\

        \rowcolor{violet!10} MedSeg-R (ours) & \textcolor{ForestGreen}{\ding{51}} & \textcolor{ForestGreen}{\ding{51}} & \textcolor{ForestGreen}{\ding{51}} \\

        \bottomrule
    \end{tabular}
    \vspace{-5mm}
    \label{tab:methods_comparison}
\end{table*}


In this paper, we introduce a novel task, medical image reasoning segmentation, which generates segmentation masks based on complex and implicit medical instructions. To achieve this, we propose MedSeg-R, an end-to-end framework that aims to enhance medical image segmentation models with the advanced reasoning abilities of MLLMs (see Fig.~\ref{fig1}(c)). Specifically, MedSeg-R includes two key components: (1) a global context understanding module that processes and interprets complex image-text instruction pairs, and (2) a pixel-level grounding module that generates precise segmentation masks by decoding comprehensive text response after multi-round reasoning. Notably, MedSeg-R is capable of comprehending implicit and complex medical instructions and autonomously producing corresponding segmentation masks. This capability streamlines the diagnostic and segmentation process, significantly improving efficiency and accuracy in medical image analysis. To further enhance MedSeg-R's reasoning segmentation capabilities and support the broader medical community, we introduce MedSeg-QA, an image-mask-conversation dataset specifically designed for medical image reasoning segmentation. MedSeg-QA is constructed through a three-stage automatic annotation pipeline, supplemented by physician-reviewed annotations to ensure high quality. The dataset comprises over 10,000 image-mask pairs, each paired with detailed multi-round conversations that comprehensively describe the medical image content. This rich dataset aims to advance the development of reasoning segmentation models in the medical domain.

\section{Method}

In this section, we first define the medical image reasoning segmentation task in Sec. \ref{part:definition}, then we first detail the architecture and training objectives of MedSeg-R in Sec. \ref{part:architecture}, followed by a description of the three-stage generation pipeline for our MedSeg-QA dataset in Sec. \ref{part:datapipeline}.


\subsection{Medical Image Reasoning Segmentation}
\label{part:definition}
Given the an medical image $\mathbf{x}_v$ and a text instruction $\mathbf{x}_l$, medical image reasoning segmentation task aims to generate comprehensive text response $\mathbf{y}$ that adheres to the instruction, while simultaneously generate precise segmentation mask $\mathbf{m}$ corresponding to the response. 

\begin{figure}[t]

    \centering
    \includegraphics[width=1.0\linewidth]{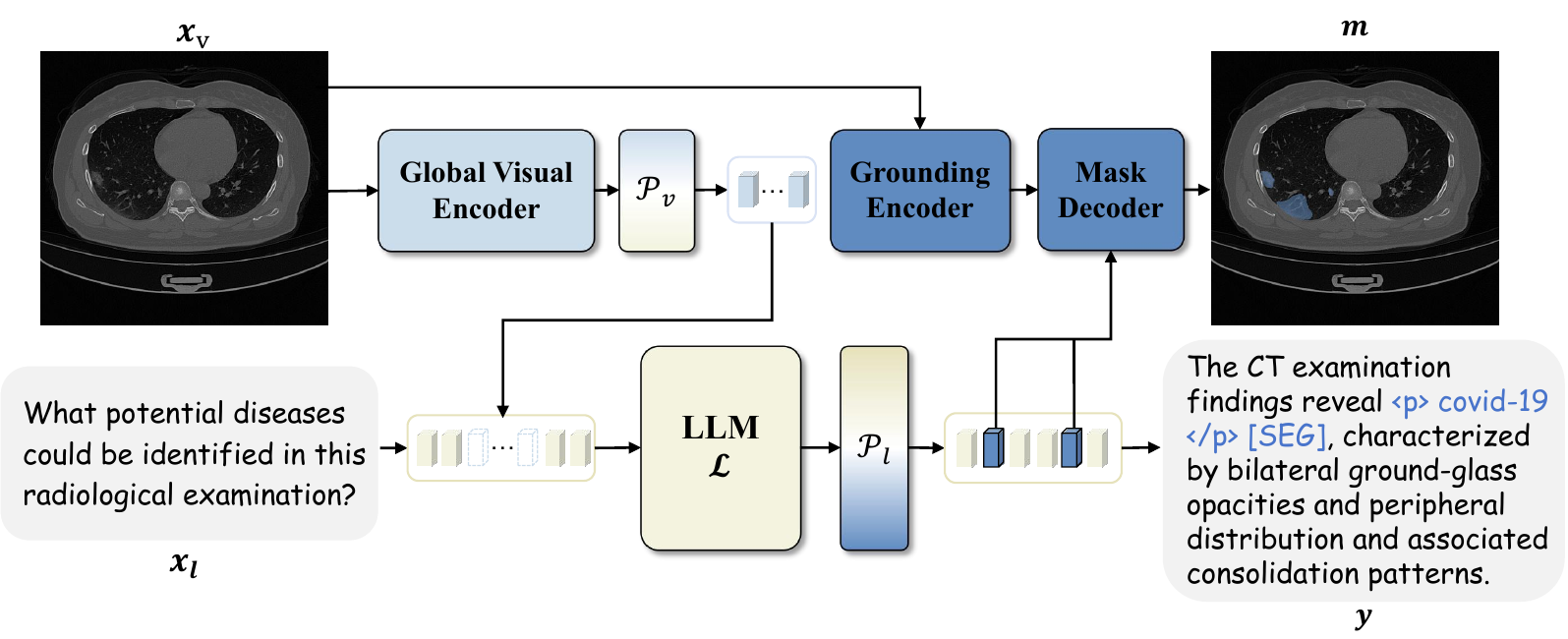}
    \vspace{-5mm}
    \caption{\textbf{MedSeg-R's architecture.} This figure illustrates our model's ability of utilizing the reasoning capabilities of the large language model (LLM) to produce detailed text responses $\mathbf{y}$ and guide the mask decoder in generating precise segmentation masks $\mathbf{m}$.}
    \vspace{-3mm}
    \label{architecture}
\end{figure}

\subsection{MedSeg-R Architecture}
\label{part:architecture}
To address the medical image reasoning segmentation task, we propose a comprehensive architecture for MedSeg-R, which consists of two primary modules: a Global Context Understanding (GCU) module and a Pixel-level Grounding (PG) module. The GCU module processes the medical image $\mathbf{x}_v$ and the text instruction $\mathbf{x}_l$ to generate a sequence of intermediate multi-modal tokens $\mathbf{h}_l$. These tokens are then fed into a text projection layer $\mathcal{P}_l$ to produce the text response $\mathbf{y}$, which contains special tokens $\mathbf{t}_{seg}$. These tokens are subsequently used to guide the PG module in generating the corresponding segmentation mask.


\noindent \textbf{Global Context Understanding Module.}
The GCU module consists of two core components: a global visual encoder ($\mathcal{V}$), realized by the CLIP~\cite{clip} image encoder with ViT-H/14~\cite{dosovitskiy2021imageworth16x16words}, and a large language model (LLM) ($\mathcal{L}$), realized by the Mistral-7B~\cite{jiang2023mistral7b} model. Given the image $\mathbf{x}_v$ and text instruction $\mathbf{x}_l$, the image is first encoded into a series of image tokens $\mathbf{s}_{v} = \mathcal{V}(\mathbf{x}_v)$. We then employ a trainable linear projection $\mathcal{P}_v$ to map the image tokens into the text embedding space, resulting in $\bm{\alpha}_{v} = \mathcal{P}_v(\mathbf{s}_{v})$. The LLM combines both the image tokens and the text instruction to generate the intermediate multi-modal tokens $\mathbf{h}_l$:
\[
    \mathbf{h}_l = \mathcal{L}([\bm{\alpha}_{v}, \mathbf{x}_l]),
\]
where $\mathbf{h}_l$ provides a global contextual understanding of the medical image $\mathbf{x}_v$, effectively integrating visual and textual information. 

\noindent \textbf{pixel-level grounding Module.}
The pixel-level grounding module adopts a SAM-like architecture, comprising a grounding encoder $\mathcal{G}$, a text projection layer $\mathcal{P}_l$, and a mask decoder $\mathcal{M}$. The grounding encoder can be implemented using common segmentation backbones, such as SAM~\cite{sam} or Mask2Former~\cite{mask2former}. In this work, we apply the SAM encoder with a ViT-H version to process the image $\mathbf{x}_v$. Given the intermediate multi-modal output $\mathbf{h}_l$ from the GCU, $\mathbf{h}_l$ is projected through $\mathcal{P}_l$, yielding the text response $\mathbf{y} = \mathcal{P}_l(\mathbf{h}_l)$. Subsequently, $\mathbf{y}$ would include special tokens $\mathbf{t}_{seg}$ 
\footnote{The special tokens \(\mathbf{t}_{seg}\) adhere to the format \texttt{<p> ... </p> [SEG]}, where the tokens enclosed within the \texttt{<p>} and \texttt{</p>} delimiters are extracted and subsequently utilized by the mask decoder to generate the corresponding segmentation mask.}, e.g., ``\texttt{<p> covid-19 </p> [SEG]} '', which are fed into the mask decoder to guide the generation of the segmentation mask $\mathbf{m}$. This process can be summarized by the following equation:
\[
    \mathbf{m} = \mathcal{M}(\mathcal{G}(\mathbf{x}_v), \mathbf{t}_{seg}).
\]

\noindent \textbf{Training Objectives.}
Following prior works~\cite{lai2023lisa,glamm}, the overall objective $L$ is a weighted sum of two losses $L_t$ and $L_m$, controlled by $\lambda_t$ and $\lambda_m$, as defined by the following equation:
\[
L = \lambda_t L_t + \lambda_m L_m,
\]
where $L_t$ represents the auto-regressive cross-entropy loss, which ensures high-quality text generation and $L_m$ combines binary cross-entropy loss and DICE loss to ensure the generation of precise segmentation masks.


\subsection{MedSeg-QA Generation Pipeline}
\label{part:datapipeline}
Currently, well-annotated medical image datasets are generally categorized into two types: i) datasets containing detailed diagnostic reports, captions, or multi-round question-answering paired with medical images but lacking segmentation annotations, and ii) datasets designed for medical image segmentation tasks, which include precise segmentation annotations but lack corresponding text descriptions. Recognizing the lack of benchmarks for the new medical image reasoning segmentation task, we introduce MedSeg-QA, a dataset comprising over 10,000 images with precise masks and comprehensive conversations describing diagnoses and image details. MedSeg-QA includes medical images from various modalities, such as CT, histological imaging, and optical imaging, covering a wide range of anatomical structures and disease types, including lung nodules, tumors, dermoscopy, and pathological slides. The dataset is generated through a three-stage pipeline: 1) image caption generation, 2) image caption refinement, and 3) structured conversation generation.


\noindent \textbf{Image Caption Generation.}
The goal of this stage is to equip existing medical segmentation datasets~\cite{covid19,kucs2024medsegbench,decathlon,shi2022ebhisegnovelenteroscopebiopsy} with initial medical image captions. To achieve this, we employ a state-of-the-art MLLM, GPT-4~\cite{gpt4}, to generate the initial captions. Specifically, we design dataset-specific prefixes to ensure the captions are contextually appropriate and highly relevant to each dataset. For example, for the COVID-19 CT dataset~\cite{covid19}, the dataset-specific prefixes are as follows: ``\texttt{Imagine you are a professional AI chest CT imaging assistant. The doctor needs to diagnose COVID-19, and you are tasked with analyzing the image to provide detailed, effective, and accurate diagnostic advice.}'' We then send these dataset-specific prefixes, along with standardized prompts and images, to GPT-4 to generate the initial image captions.


\noindent \textbf{Image Caption Refinement.}
In this stage, we refine the initial image captions through a thorough review by physicians to assess their accuracy. Images with correct captions are retained, while those with incorrect captions are resubmitted to GPT-4 for a second round of generation with adjusted prompts. If the second round fails to produce satisfactory results, the images are manually annotated by physicians, ensuring the refined captions are accurate and reliable.

\noindent \textbf{Conversations Generation.}
Finally, we generate multi-round conversations based on the refined captions to enhance reasoning capabilities. Specifically, we provide GPT-4 with a pre-defined question list and refined captions, prompting it to generate appropriate conversations while highlighting segmentation elements in the answer using the special tokens [SEG] described in Sec.~\ref{part:architecture}.




\vspace{-5mm}
\section{Results and Analyses}


\begin{figure}[t]
    \centering
    \includegraphics[width=1.0\linewidth]{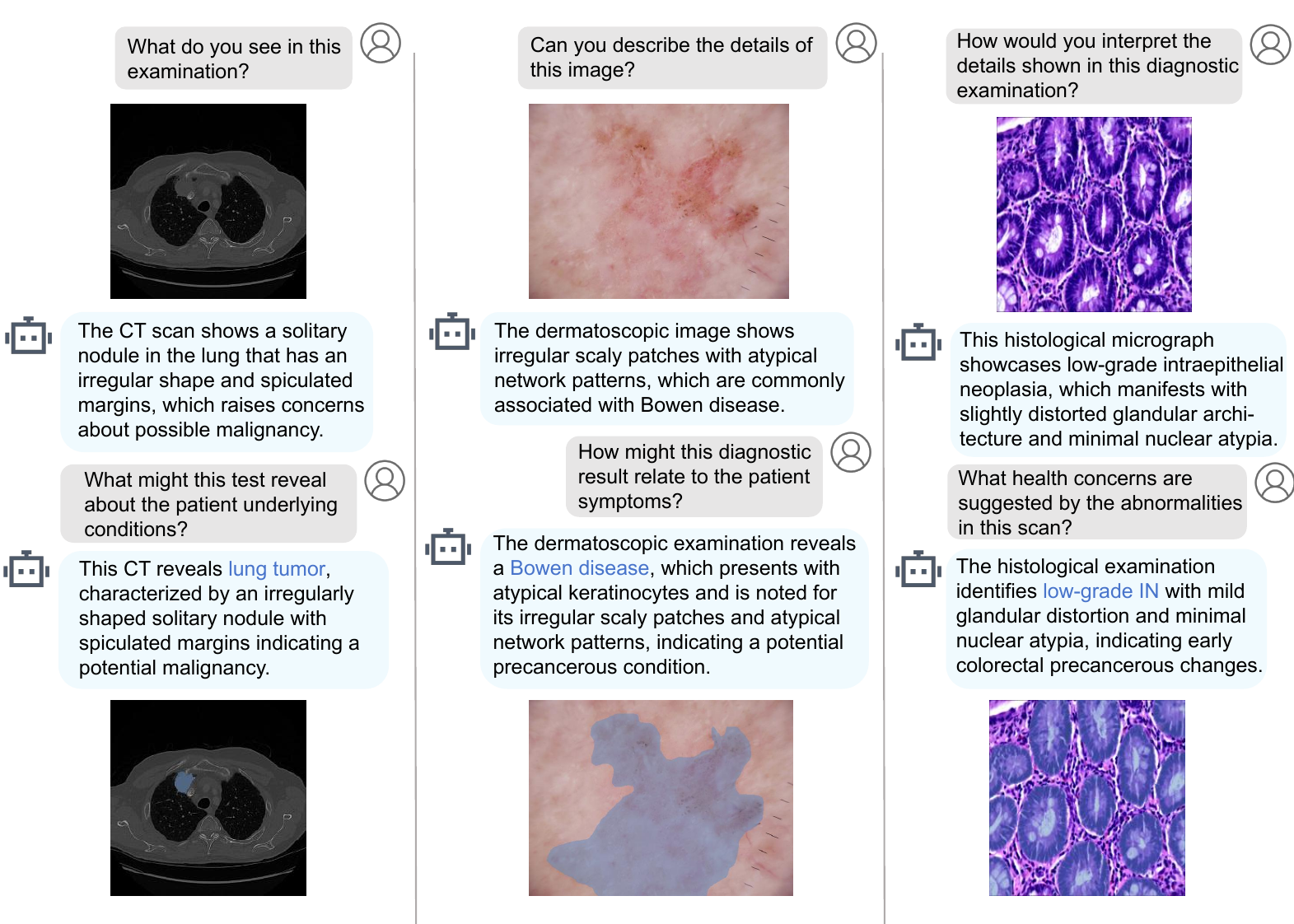}
    \caption{\textbf{Qualitative evaluation of MedSeg-R on the MedSeg-QA dataset.} The figure illustrates examples of MedSeg-R processing complex human instructions and provide corresponding segmentation masks across various modalities of medical images.}
    \vspace{-8mm}
    \label{fig:conversation}
\end{figure}

\begin{table}[!t]
\centering
\small 
\setlength{\tabcolsep}{3pt} 
\renewcommand{\arraystretch}{1.2} 

\caption{Dataset statistics for medical VQA benchmarks. The values represent the number of samples in each subset. For SLAKE, we only consider the English subset.}
\label{tab:data_stat_three_datasets}

\begin{tabular}{l c c | c c c | c c c}  
\toprule
\multirow{2}{*}{\textbf{Metric}} & 
\multicolumn{2}{c|}{\textbf{VQA-RAD}~\cite{vqa_rad}} & 
\multicolumn{3}{c|}{\textbf{SLAKE}\cite{slake}} & 
\multicolumn{3}{c}{\textbf{PathVQA}\cite{pathvqa}} \\ 
\cmidrule(lr){2-3} \cmidrule(lr){4-6} \cmidrule(lr){7-9}
& Train & Test & Train & Val & Test & Train & Val & Test \\
\midrule
\textbf{Images}       & 313   & 203   & 450   & 96   & 96    & 2,599  & 858   & 858   \\
\textbf{QA Pairs}     & 1,797 & 451   & 4,919 & 1,053 & 1,061 & 19,755 & 6,279 & 6,761 \\
\cmidrule(r){1-9}
\textbf{Open}   & 770   & 179   & 2,976 & 631   & 645   & 9,949  & 3,144 & 3,370 \\
\textbf{Closed} & 1,027 & 272   & 1,943 & 422   & 416   & 9,806  & 3,135 & 3,391 \\
\bottomrule
\end{tabular}
\vspace{-4mm}
\end{table}

\subsection{Qualitative Evaluation on the MedSeg-QA Dataset}
To demonstrate the reasoning-based segmentation capability of MedSeg-R, we divide the MedSeg-QA dataset into training and validation sets. Several representative examples from the validation set are illustrated in Fig.~\ref{fig:conversation}. 

As observed, despite the implicit nature of the user's questions, MedSeg-R is capable of generating detailed and context-aware descriptions of the medical images. Furthermore, it accurately identifies the corresponding abnormal regions and produces precise segmentation masks, which highlight its ability to integrate reasoning with pixel-level grounding effectively.

\subsection{Evaluation on Biomedical VQA and Segmentation Tasks}  
To demonstrate that MedSeg-R is also effective in standard biomedical visual question answering (VQA) and segmentation tasks, we conduct experiments in both tasks and achieve promising results.

\noindent \textbf{Comparison with SoTA on Biomedical VQA Task}  
We evaluate our model on three widely used biomedical VQA datasets, with dataset details summarized in Table \ref{tab:data_stat_three_datasets}. For closed-ended questions, we report accuracy as the evaluation metric, assessing the model’s ability to correctly classify predefined answers. For open-ended questions, we employ recall, which measures the proportion of ground-truth tokens appearing in the generated responses, ensuring a fair evaluation of the model’s language generation capabilities in medical VQA.
Results are shown in Table \ref{tab:sota_vqa}.

\begin{table*}[!t]
    \centering
    \caption{Comparison with prior state-of-the-art supervised methods. We present results for both open-ended and closed-form question answering across three datasets. The ``--'' symbol denotes results that are not available.}
    \renewcommand{\arraystretch}{1.2} 
    \setlength{\tabcolsep}{4pt} 
    \begin{tabular}{l|cc|cc|cc}  
        \toprule
        \multirow{2}{*}{\textbf{Method}} & \multicolumn{2}{c|}{\textbf{VQA-RAD}} & \multicolumn{2}{c|}{\textbf{SLAKE}} & \multicolumn{2}{c}{\textbf{PathVQA}}  \\
        \cmidrule(lr){2-3} \cmidrule(lr){4-5} \cmidrule(lr){6-7}
        & Open  & Closed  & Open  & Closed  & Open  & Closed  \\ 
        \midrule
        VL Encoder–Decoder~\cite{bazi2023vision}  & 71.49  & 82.47  & --     & --     & 71.49  & 85.61  \\
        Prefix T. Medical LM~\cite{van2023open}  & --     & --     & 84.30  & 82.01  & 40.00  & 87.00  \\ 
        PubMedCLIP~\cite{pubmedclip}   & 60.10  & 80.00  & 78.40  & 82.50  & --     & --     \\
        BiomedCLIP~\cite{zhang2023biomedclip}  & 67.60  & 79.80  & 82.05  & 89.70  & --     & --     \\
        LLaVA-Med~\cite{llavamed}  & 64.75  & 83.09  & \textbf{87.11} & 86.78  & 39.60  & 91.09   \\
       \rowcolor{LGray} MedSeg-R (ours)   & \textbf{72.90}  & \textbf{84.45}  & 84.62  & \textbf{91.30}  & \textbf{72.83}   &  \textbf{91.64}\\
        \bottomrule
    \end{tabular}  
    \vspace{-4mm}
    \label{tab:sota_vqa}
\end{table*}

\noindent \textbf{Comparison with SoTA on Medical Image Segmentation Task} 
To evaluate the performance of our model in medical image segmentation, we conducted a comparative analysis against several frameworks commonly employed in medical image segmentation competitions. The evaluation was performed on the FLARE 2022 dataset, which comprises CT scans featuring 13 distinct abdominal organs, using two key metrics: the Dice Similarity Coefficient (DSC) and the Normalized Surface Distance (NSD).


To ensure consistency and fairness in the evaluation process, we adopted a simple and uniform text instruction to guide the model in generating accurate segmentation masks. The instruction template used was:``\texttt{Please segment the <class-name> in the medical image,}'' where ``\texttt{<class-name>}'' represents one of the 13 abdominal organ names in the dataset. The model was designed to respond with: ``\texttt{Sure, it is [SEG].}'' In this setup, the ``\texttt{[SEG]}'' token serves as the special token to enable the mask decoder to produce the corresponding segmentation mask. 

As demonstrated by the results in Table \ref{tab:ct_performance}, our model maintains competitive segmentation performance compared to current state-of-the-art methods.

\begin{table}[htbp]
\centering
\setlength{\tabcolsep}{12pt} 
\vspace{-2mm}
\caption{Performance comparison of different methods on FLARE 2022 dataset, evaluated using DSC and NSD.}
\begin{tabular}{lcc}
\toprule
\multirow{2}{*}{\textbf{Methods}} & \multicolumn{2}{c}{\textbf{Organs in FLARE 2022}} \\
\cmidrule(lr){2-3}
 & \textbf{DSC} & \textbf{NSD} \\
\midrule
nnU-Net~\cite{nnunet}     & 0.8615 ± 0.0790  & 0.8972 ± 0.0824 \\
SegResNet~\cite{myronenko20193d}  & 0.7927 ± 0.1162  & 0.8257 ± 0.1194 \\
UNETR~\cite{hatamizadeh2022unetr}        & 0.6824 ± 0.1506  & 0.7004 ± 0.1577 \\
SwinUNETR~\cite{hatamizadeh2021swin}   & 0.7594 ± 0.1095  & 0.7663 ± 0.1190 \\
MedSeg-R (ours)   & \textbf{0.8718 ± 0.1034}  & \textbf{0.9091 ± 0.0917} \\
\bottomrule
\end{tabular}
\vspace{-2mm}
\label{tab:ct_performance}
\end{table}

\vspace{-8mm}
\section{Conclusion}
In this study, we propose MedSeg-R, an end-to-end framework integrating multimodal large language models to enhance medical image segmentation with advanced reasoning. We also introduce MedSeg-QA, a large-scale dataset with over 10,000 physician-validated annotations. Experiments show MedSeg-R surpasses existing models in segmentation accuracy and reasoning, achieving state-of-the-art results. By combining medical reasoning with pixel-level precision, MedSeg-R advances intelligent medical image analysis, improving clinical diagnostics and real-world applications.

\clearpage
{
\bibliographystyle{splncs04}
\bibliography{Medseg}
}
\end{document}